\documentclass{frontiersSCNS} 

\usepackage{url,hyperref,lineno,microtype,graphicx}
\usepackage{subcaption}
\usepackage[onehalfspacing]{setspace}
\usepackage{amsmath}

\def\keyFont{\fontsize{8}{11}\helveticabold }
\def\firstAuthorLast{McClure {et~al.}} 
\def\Authors{Patrick McClure\,$^{1,*}$ and Nikolaus Kriegeskorte\,$^{1}$}
\def\Address{$^{1}$MRC Cognition and Brain Science Unit, Cambridge, UK}
\def\corrAuthor{Corresponding Author}
\def\corrEmail{Patrick.McClure@mrc-cbu.cam.ac.uk}

\begin{document}
\onecolumn
\firstpage{1}

\title[Representational Distance Learning]{Representational Distance Learning for Deep Neural Networks} 

\author[\firstAuthorLast ]{\Authors} 
\address{} 
\correspondance{} 

\extraAuth{}

\maketitle

\begin{abstract}

\section{}

Deep neural networks (DNNs) provide useful models of visual representational transformations. We present a method that enables a DNN (student) to learn from the internal representational spaces of a reference model (teacher), which could be another DNN or, in the future, a biological brain. Representational spaces of the student and the teacher are characterized by representational distance matrices (RDMs). We propose representational distance learning (RDL), a stochastic gradient descent method that drives the RDMs of the student to approximate the RDMs of the teacher. We demonstrate that RDL is competitive with other transfer learning techniques for two publicly available benchmark computer vision datasets (MNIST and CIFAR-100), while allowing for architectural differences between student and teacher. By pulling the student's RDMs towards those of the teacher, RDL significantly improved visual classification performance when compared to baseline networks that did not use transfer learning. In the future, RDL may enable combined supervised training of deep neural networks using task constraints (e.g. images and category labels) and constraints from brain-activity measurements, so as to build models that replicate the internal representational spaces of biological brains.

\tiny
 \keyFont{ \section{Keywords:} neural networks, transfer learning, distance matrices, visual perception, computational neuroscience} 
\end{abstract}

\section{Introduction}

Deep neural networks (DNNs) have recently been highly successful for machine perception, particularly in the areas of computer vision using convolutional neural networks (CNNs) \citep{krizhevsky2012imagenet} and speech recognition using recurrent neural networks (RNNs) \citep{deng2013new}. The success of these methods depends on their ability to learn good, hierarchical representations for these tasks \citep{bengio2012deep}. DNNs have not only been useful in achieving engineering goals, but also as models of computations in biological brains. Several studies have shown that DNNs trained only to perform object recognition learn representations that are similar to those found in the human ventral stream \citep{khaligh2014deep,yamins2014performance,gucclu2014deep}. The models benefit from task training, which helps determine the large number of parameters and bring the domain knowledge required for feats of intelligence such as object recognition into the models. This is in contrast to the earlier approach in visual computational neuroscience of using nonlinear systems identification techniques to set the parameters exclusively on the basis of measured neural responses to large sets of stimuli \citep{naselaris2011encoding}. The latter approach is challenging for deep neural networks, because the high cost of brain-activity measurement limits the amount of data that can be acquired \citep{yamins2016using}. Ultimately, task-based constraints will have to be combined with constraints from brain-activity measurements to model information processing in biological brains.

Here we propose a method that enables the training of DNNs with combined constraints on the desired outputs and the internal representations. We demonstrate the method by using another neural net model as the reference system whose internal representations the DNN is to emulate. One method for doing so would be to have a layer in a DNN linearly predict individual measured responses (e.g. fMRI voxels or neurons), and backpropagate the error derivatives from the linear measured-response predictors into the DNN. However, the linear measurement prediction model has a large number of parameters ($n_{units} \times n_{responses}$). An alternative approach is to constrain the DNN to replicate the representational distance matrices (RDMs) estimated from brain responses. In this paper, we take a step in that direction by considering the problem of training a DNN (student) to model the sequence of representational transformations in another artificial system (teacher), a CNN trained on different data.

Our technique falls in the class of transfer learning methods. In the deep learning literature, several such techniques have been proposed both for pulling a DNN's internal representations towards the task target and for transferring knowledge from a teacher DNN to a student DNN. We begin by briefly considering the previous transfer learning approaches.

\subsubsection*{Auxiliary Classifiers: Pulling internal representations toward the desired output...} 

Recently, it has been investigated how the error signal reaching an internal layer through backpropagation can be complemented by auxiliary error functions. These more directly constrain internal representations using auxiliary optimization goals. A variety of methods using auxiliary error functions to pull representations toward the desired output have been proposed

\cite{weston2012deep} proposed semi-supervised embeddings to augment the error from the output layer. A reference embedding of the inputs was used to guide representational learning. The embedding constraint was implemented in different ways: inside the network as a layer, as part of the output layer, or as an auxiliary error function that directly affected a particular hidden layer. Weston et al. discussed a variety of embedding methods that could be used, including multidimensional scaling (MDS) \citep{kruskal1964multidimensional} and Laplacian Eigenmaps \citep{belkin2003laplacian}. The addition of these semi-supervised error functions led to increased accuracy compared to DNNs trained using output layer backpropagation alone.

\cite{lee2014deeply} also showed that auxiliary error functions improve DNN representat==ional learning. Instead of using semi-supervised methods, they performed classification with a softmax or L2SVM readout at a given intermediate hidden layer. The softmax layer allowed the output of a network to be treated as a probability distribution by performing normalized exponentiation on the previous layer's activations ($y_i = e^{x_i}/\sum_{j} e^{x_j})$. The error of the intermediate-level readout was then backpropagated to earlier layers to drive intermediate layers directly towards the target output. The gradients from these classifiers were linearly combined with the gradients from the output layer classifier. This technique resulted in improved accuracies for several datasets.

A challenge in training very deep networks is the problem of vanishing gradients. Layers far from the output may receive only a weak learning signal via conventional backpropagation. Auxiliary error functions were successfully applied to these very deep networks by \cite{szegedy2014going} to inject a complementary learning signal at internal layers by constraining representations to better discriminate between classes.  This was implemented in a very large CNN which won the ILSVRC14 classification competition \citep{russakovsky2014imagenet}. In this DNN, two auxiliary networks were used to directly backpropagate from two intermediate layers back through the main network. Similar to the method used in \cite{lee2014deeply}, the parameters for the layers in the main network directly connected to auxiliary networks were updated using a linear combination of the backpropagated gradients from later layers and the auxiliary network.

\cite{wang2015training} investigated the effectiveness of auxiliary error functions in very large CNNs and their optimal placement. They selected where to place these auxiliary functions by measuring the average magnitude of the conventional backpropagation error signal at each layer. Auxiliary networks, similar to those used in \cite{szegedy2014going}, were placed after layers with vanishing gradients. These networks consisted of a convolutional layer followed by three fully connected layers and a softmax classifier. As in \cite{lee2014deeply} and \cite{szegedy2014going}, the auxiliary gradients were linearly combined to update the model parameters. Adding these supervised auxiliary error functions led to an improved accuracy for two very large datasets, ILSVRC12 \citep{russakovsky2014imagenet} and MIT Places \citep{zhou2014learning}.

\subsubsection*{Transfer Learning: Pulling the representations of a student towards those of a teacher...}
Enabling a student network to learn from a teacher is useful for a number of tasks, for instance model compression (also known as knowledge distillation) and transfer learning \citep{bengio2012deep}. The goal in either case is to use the representational knowledge learned by a teacher neural network to improve the performance of a student network \citep{bucilua2006model,ba2014deep,hinton2015distilling}. For model compression, the teacher is a larger or more complex network with higher performance than the student. For knowledge transfer, the representations learned by the teacher network are used to improve the training of a student network on a different tasks or using different data. Several techniques have been proposed for performing these methods.

One technique for model compression is to have the student learn the output representation of the teacher for a given training input. For classification, the neurons before the softmax layer can be constrained to have the same values as the teacher using mean squared error (MSE) as done in \cite{bucilua2006model,ba2014deep}. Alternatively, the output of the softmax layer can be constrained to represent the same, or similar, output distribution as the teacher. This can be done by minimizing the cross-entropy between the output distributions of the teacher and student networks for the training inputs \citep{hinton2015distilling}. However, these techniques assume that the student is learning the same task as the teacher.

Knowledge from different networks can also be transferred at internal layers. \cite{romero2014fitnets} proposed a method for transferring the knowledge of a wide and shallow teacher to a thin and deep student, called FitNet.  Pre-trained a network by constraining an intermediate layer of the student network to have representations that could linearly predict 'hints' from the teacher network (i.e. activation patterns at a corresponding layer in the teacher network). After this, the network was fine-tuined using the technique proposed in \cite{hinton2015distilling}. The FitNet method was shown to improve the student’s classification accuracy.

Another prominent technique for performing transfer learning is to initialize the weights of the student network to those of the teacher. The network is then trained on a different task or using different data. This can lead to improved network performance \citep{yosinski2014transferable}. However, this requires that the teacher and student have the same, or very similar, architectures, which may not be desirable, especially if the teacher is a biological neural network.

In this paper, we introduce an auxiliary error function that enables a student network to learn from the internal representational spaces of a teacher that has a similar or  different architecture. The method constrains the student's representational distances in a set of layers to approximate those of the teacher. The student can thus learn the computational transformations discovered by the teacher, leading to improved representational learning during training.

\section{Methods}

Our method, representational distance learning (RDL), enables DNNs to learn from the representations of other models to improve performance. As in \citet{lee2014deeply,szegedy2014going,wang2015training}, we utilize auxiliary error functions to train internal layers directly in conjunction with the error from the output layer found via backpropagation. We propose an error function that maximizes the similarity between the representational spaces of a student DNN and that of a teacher model.

\begin{figure}[!htb]
\centering
\includegraphics[scale=.275]{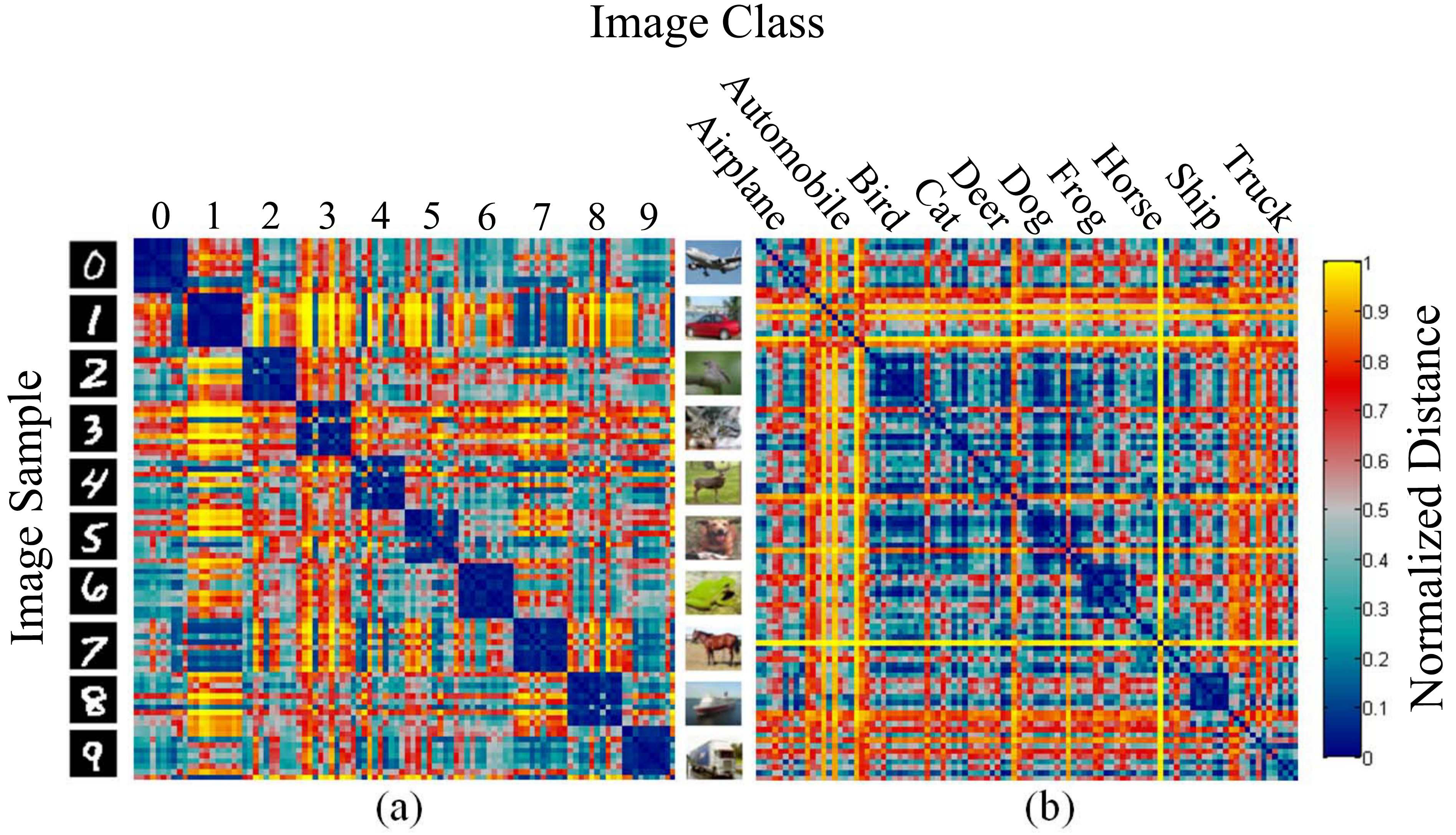}
\caption{Example representational distance matrices (RDMs) of the output layer of convolutional neural networks (CNNs) for ten random images of each class from (a) MNIST and (b) CIFAR-10 made using the RSA toolbox \citep{nili2014toolbox}.}
\label{fig:RDMs}
\end{figure}

\subsection{Representational Distance Matrices}

In order to compare the representational spaces of models, a method must be used to describe them. As discussed in \cite{weston2012deep}, a representational space can be characterized by the pairwise distances between representations. This idea has been used in several methods such as MDS, which seeks to reduce the dimensionality of data while minimizing the error between the pairwise distance matrix of the original data and the reduced dimensionality data \citep{kruskal1964multidimensional}. 

\cite{kriegeskorte2008representational} proposed using the matrix of pairwise dissimilarities between representations of different inputs, which they called representational distance, or dissimilarity, matrices (RDMs), to compare computational models and neurological data. More recently, \cite{khaligh2014deep} used this technique to analyze several computer vision models, including the CNN proposed in \cite{krizhevsky2012imagenet}, and neurological data. Any distance function could be used to compute the pairwise dissimilarities, for instance the Euclidean or correlation distances. An RDM for a DNN can be defined by:

\begin{small}
\begin{equation}
RDM(X;f_m)_{i,j} = d(f_m(x_i;W_m),f_m(x_j;W_m))
\end{equation}
\end{small}

\noindent
where $X$ is a set of $n$ inputs (e.g. a mini-batch or a subset of a mini-batch), $f_m$ is the neuron activations at layer $m$, $x_i$ and $x_j$ are single inputs, $W_m$ is the weights of the neural network up to layer $m$, and some distance, or dissimilarity, measure $d$. 

In addition to characterizing the information present in a particular layer of a DNN, RDMs can be used to visualize the representational space of a layer in a DNN (Figure \ref{fig:RDMs}). Currently, understanding and visualizing the information captured by internal layers in a DNN is challenging. \citet{zeiler2014visualizing} proposed a method for visualizing the input features which active internal neurons at varying layers using deconvolutional neural networks. \citet{yosinski2015understanding} also proposed methods for visualizing the activations of a DNNs for a given input. However, these methods do not show the categorical information of each representational layer. Visualizing the similarity of labelled inputs at layers of interest, via an RDM, allow clusters inherent to the learned representational transformations to be viewed.

\subsection{Representational Distance Learning}

RDL uses an auxiliary error functions that maximizes the similarity between the RDMs of a student and the RDMs of a teacher at several layers. This is motivated by the idea that RDMs, or distance matrices in general, can characterize the representational space of a model. DNNs seek to learn a set of hierarchical representations. For classification, this culminates in finding a representational space where different classes are separable. RDL allows a DNN to learn from the representations of a different, potentially better, model by maximizing the similarity between the RDMs of the DNN being trained and the target model at several layers. Unlike in \citet{bucilua2006model,ba2014deep,hinton2015distilling}. RDL not only directly trains the output representation, but also the representations of hidden layers. As discussed in \citet{bengio2012deep}, however, large datasets can prohibit the use of pairwise techniques, since the number of comparisons grows quadratically with dataset size. To address this, our technique only uses a random subset of all pairwise distances for each parameter update. This allows the speed of our method to be constrained by the subset size and not the overall number of training examples, which is usually several orders of magnitude larger.

In order to maximize the similarity between the RDM of a DNN layer being trained and a target RDM, we propose minimizing the mean squared error between the two RDMs. This corresponds to making all possible pairwise distances as similar as possible:

\begin{small}
\begin{equation}
E_{aux}(X;f_m;T_m)=\frac{2}{n(n-1)}\sum_{(i,j)|i<j} (RDM(X;f_m)_{i,j}-T_{m,i,j})^2
\end{equation}
\end{small}

\noindent
where $X$ is a set of $n$ inputs (e.g. a mini-batch or a subset of a mini-batch), $f_m$ is the neuron activations at layer $m$, and $T_{m,i,j}$ is the distance between the teacher's representations of input $x_i$ and input $x_j$ at layer $m$. The function $d$ used to calculate the RDMs (Eq. 1) could be any dissimilarity or distance function, but we chose to use the mean squared error (MSE). This results in the average auxiliary error with respect to neuron $k$ of $f_m$ , $f_{m,k}$, for input $x_i$ and the weights of the neural network up to layer $m$, $W_m$, being defined as:

\begin{small}
\begin{equation}
\frac{\partial E_{aux}(x_i;X;f_m;T_m)}{\partial f_{m,k}} = \frac{8}{n(n-1)} \sum_{j | j \neq i} (RDM(X;f_m)_{i,j}-T_{m,i,j}) (f_{m,k} \rvert_{x_j}^{x_i})
\end{equation}
\end{small}

\noindent
where $f_{m,k} \rvert_{x_j}^{x_i} = f_{m,k}(x_i;W_m)-f_{m,k}(x_j;W_m)$.

However, calculating the error for every pairwise distance can be computational expensive, so we estimate the error using a random subset, $P$, of the pairwise distances for each update of a network's parameters. This leads to the auxiliary error gradient being approximated by:

\begin{small}
\begin{equation}
\frac{\partial E_{aux}(x_i;X;f_m;T_m)}{\partial f_{m,k}} \approx \frac{8}{|X_P||P_{x_i}|} \sum_{(i,j) \in P_{x_i}} (RDM(X;f_m)_{i,j}-T_{m,i,j}) (f_{m,k} \rvert_{x_j}^{x_i})
\end{equation}
\end{small} 

\noindent
where $X_P$ is the set of all images contained in $P$, $P_{x_i}$ is the set of all pairs, $(i,j)$, in $P$ that include input $x_i$ and another input, $x_j$. If an image is not sampled, its auxiliary error is zero.

The total error of $f_{m,k}$ for input $x_i$ is calculated by taking a linear combination of the auxiliary error at layer $m$ and the error from backpropagation of the output error function and any later auxiliary functions. These terms are combined using weighting hyper parameter $\alpha$, similar to the method discussed in \citet{lee2014deeply}, \citet{szegedy2014going}, and \citet{wang2015training}. In RDL, $\alpha$ is the weight of the RDL error in the overall error function. Subsequently, the error gradient at a layer with an auxiliary error function is defined as:

\begin{small}
\begin{equation}
\frac{\partial E_{total}(x_i;y_i;X;f_m;T_m)}{\partial f_{m,k}} = \frac{\partial E_{backprop}(x_i;y_i;f_m)}{\partial f_{m,k}} + \alpha \frac{\partial E_{aux}(x_i;X;f_m;T_m)}{\partial f_{m,k}}
\end{equation}
\end{small}

This error is then used to calculate the error of earlier layers in the DNN using backpropagation. As discussed by \citet{lee2014deeply} and \citet{wang2015training}, the value of $\alpha$ was decayed as training progressed. Throughout training, $\alpha$ was updated following $\alpha_{t+1}=\alpha_0*(1-t/t_{max})$ where $t$ is the epoch number and $t_{max}$ is the total number of epochs. By using this decay rule, the auxiliary error function initially helps drive the parameters to good values while allowing the DNN to converge predominantly using the output error by the end of training.

\begin{table}[b]
\caption{The convolutional neural network (CNN) architecture used for MNIST.}
\label{MNIST-arch}
\begin{center}
\begin{small}
\begin{tabular}{|c|c|c|c|c|c|}

\hline
Layer & Kernel Size & \# Features & Stride & Non-linearity & Other \\
\hline
Conv-1 & 5x5 & 32 & 1 & ReLU & - \\
\hline
MaxPool-1 & 3x3 & 32 & 3 & Max & - \\
\hline 
Conv-2 & 5x5 & 64 & 1 & ReLU & - \\
\hline
MaxPool-2 & 2x2 & 64 & 2 & Max & - \\
\hline
FC & 1500 & 200 & - & ReLU & Dropout ($p=0.5$)\\
\hline
Linear & 200 & 10 & - & - & - \\
\hline

\end{tabular}
\end{small}
\end{center}
\end{table}

\section{Results}

To evaluate the effectiveness of RDL, we perform two experiments using four different datasets, MNIST, InfiMNIST, CIFAR-10, and CIFAR-100. For each experiment, we transferred the knowledge of a teacher network trained on a separate dataset to a student network with the a similar architecture using: (1) finetuning after directly copying the weights of the teacher, (2) pre-training an internal layer of the student to linearly predict a corresponding layer in the teacher using 'hints', and (3) using RDL. We compared the results to two non-transfer learning networks, a network only constrained at the output layer using the target labels and a deeply supervised network, which constrained both the output layer and internal layers using the target labels. We implemented all of these methods using Torch \citep{collobert2011torch7}. These experiments show that the knowledge stored in the weights of a teacher network can be transferred to a student network using the representational distances learned by a teacher trained on a related task.

\begin{table}
\caption{The McNemar exact test p-values for the tested CNNs trained on MNIST. Arrows indicate a significant difference ($p < 0.05$,uncorr.) and point to the better model.}
\label{MNIST-McNemar}
\vskip 0.15in
\begin{tiny}
\begin{sc}
\centering

\begin{tabular}{|c|c|c|c|c|c|c|}
\cline{2-7}
\multicolumn{1}{c|}{ }& Baseline & Teacher & Finetuning & Deep Supervision & Hints & RDL \\
\hline
Baseline & --- & 0.38 & 0.00 $\uparrow$ & 0.11 & 0.34 & 0.01 $\uparrow$ \\
\cline{1-1}
Teacher & 0.38 & --- & 0.01 $\uparrow$ & 0.66 & 0.89 & 0.20 \\
\cline{1-1}
Finetuning &  0.00 $\leftarrow$ & 0.01 $\leftarrow$ & --- & 0.14 & 0.04 $\leftarrow$ & 0.63  \\
\cline{1-1}
Deep Supervision & 0.11 & 0.66 & 0.14 & --- & 0.64 & 0.39 \\
\cline{1-1}
Hints & 0.34 & 0.89 & 0.04 $\uparrow$ &0.64 & --- & 0.17 \\
\cline{1-1}
RDL & 0.01 $\leftarrow$ & 0.20 & 0.63 & 0.39 & 0.17 & --- \\
\hline
\end{tabular}
\end{sc}
\end{tiny}
\end{table}

\subsection{MNIST}

MNIST is a dataset of 28x28 images of handwritten digits from ten classes, 0 through 9 \citep{lecun1998gradient}. The dataset contains 50,000 training images and 10,000 test images. A 10,000 image subset of the training data was used as a validation set for hyper-parameter tuning. No pre-processing or data augmentation was applied. InfiMNIST is a dataset that extends the MNIST dataset using pseudo-random deformations and translations \citep{loosli2006training}. The first 10,000 non-MNIST InfiMNIST examples were used as a validation set and the next 120,000 examples were used as a training set for the teacher network. Each tested network had the same architecture (Table \ref{MNIST-arch}), excluding any auxiliary error functions. The deeply supervised network had linear auxiliary softmax classifiers placed after the max pooling layers and $\alpha$ was decayed using $\alpha_{t+1} = \alpha_t * 0.1 * (1-t/t_{max})$, as proposed in \citet{lee2014deeply}. For the finetuning network, the weights were initialized as the weights of the teacher network instead of being randomly initialized. After this, the network was trained normally. The RDL network had auxiliary error functions after both max pooling layers and the fully connected layer. 5\% (500) of the image pairs per mini-batch were used to calculate the RDL auxiliary errors. A momentum of 0.9 and a mini-batch size of 100 were used for all networks trained on MNIST and InfiMNIST.

\begin{figure}[b]
\centering
\includegraphics[scale=.3]{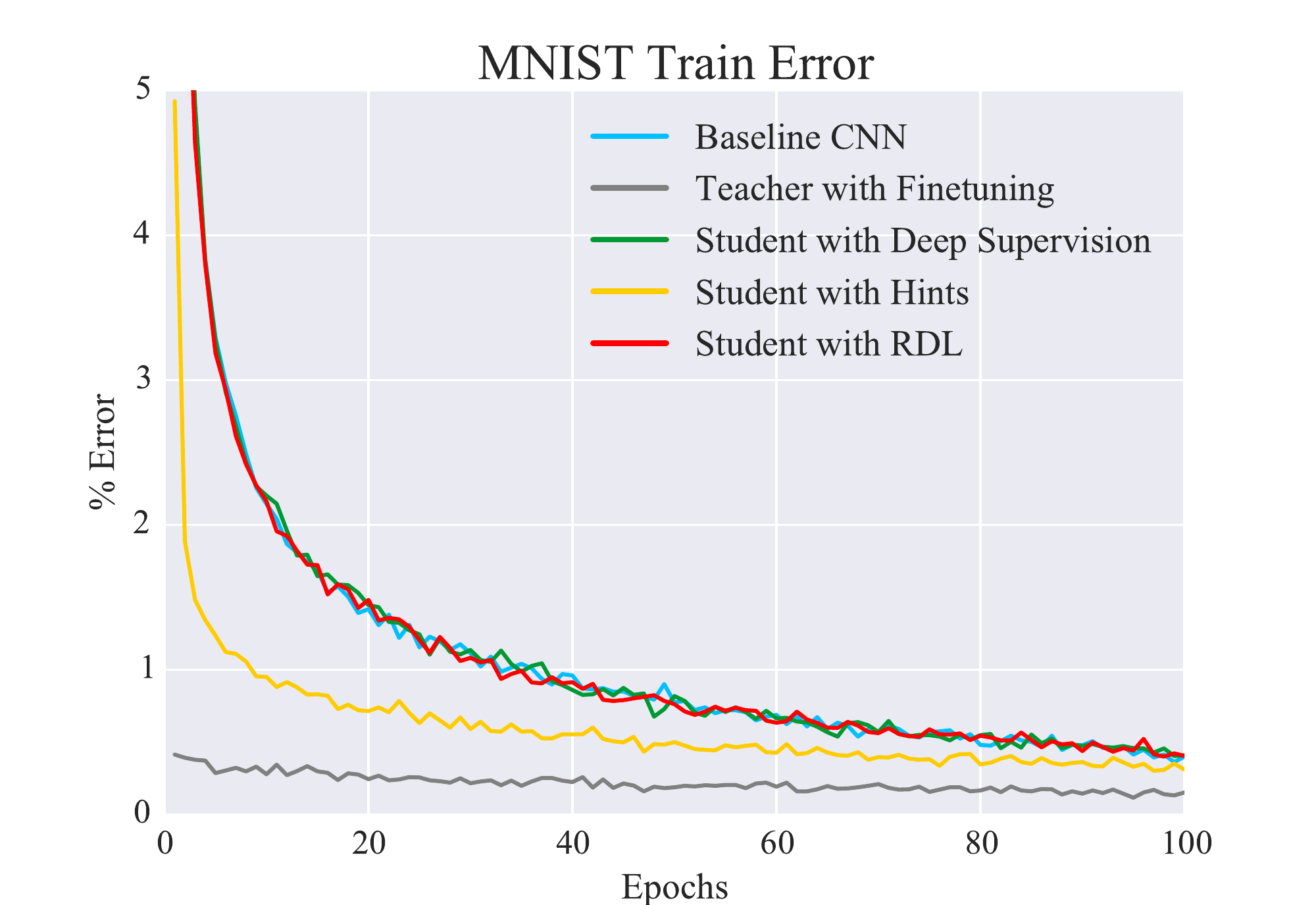}
\includegraphics[scale=.3]{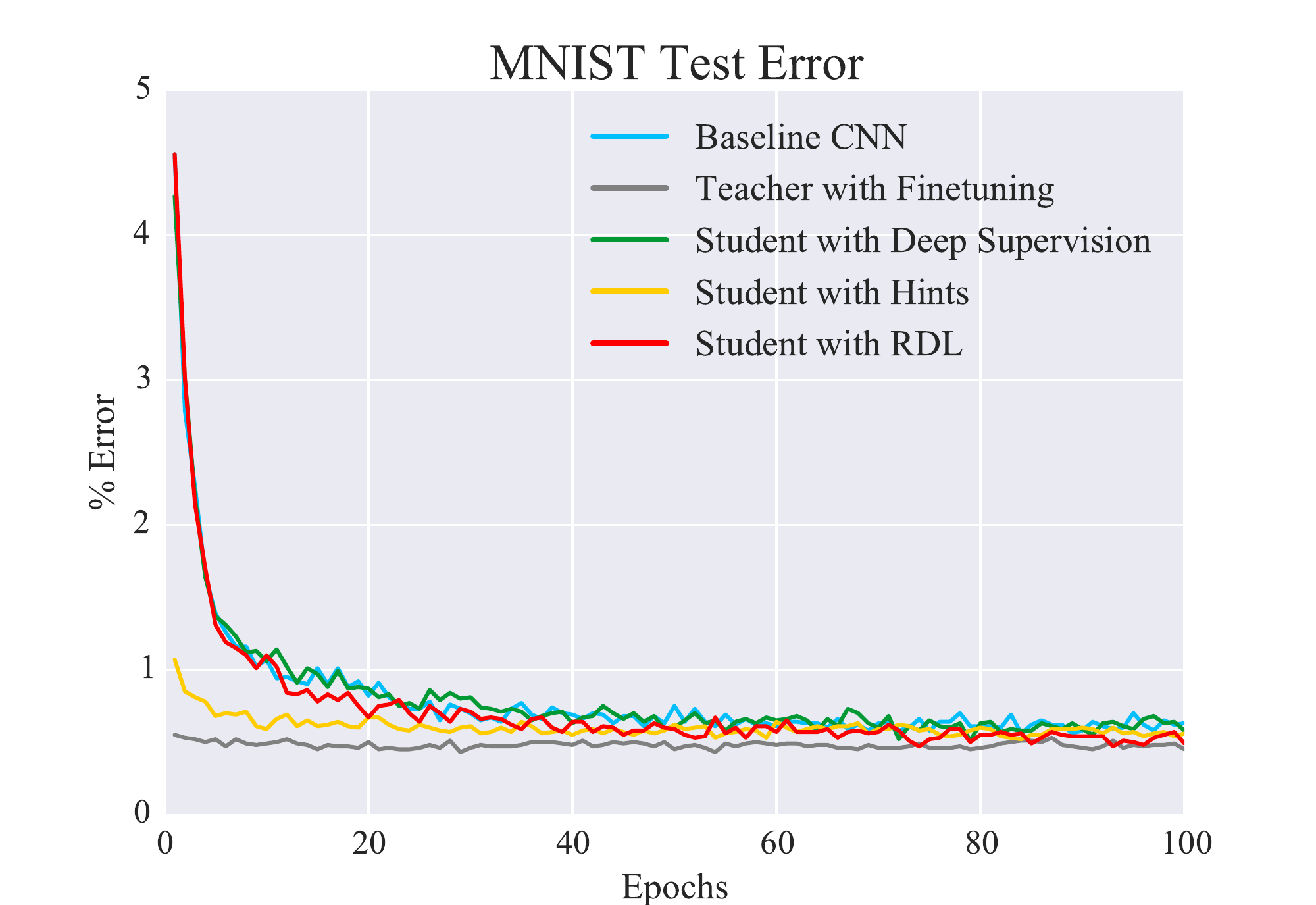}
\caption{The change in the train and test errors through time as the tested convolutional neural networks (CNNs) are trained on MNIST.}
\label{fig:MNISTTrainError}
\end{figure}

In addition to the classification error (Figure \ref{fig:MNISTTrainError} and Table \ref{Cifar-errors}), we used the McNemar exact test \citep{edwards1948note} to evaluate whether a network was significantly more accurate in classifying a random image from the distribution from which the images in the training and test sets were drawn. The results (Table \ref{MNIST-McNemar}) show that the finetuning and RDL methods both signifantly improve accuracy compared to the baseline CNN. They are, however, not significantly different, showing the ability of RDL to indirectly transfer the knowledge of the teacher network. The finetuned network is also significantly better than the teacher and the 'hint' network, unlike RDL. This is because RDL actively constrains the student network to imitate the teacher, while finetuning only affects initialization.

\begin{figure}
\centering
\graphicspath{MNIST_RDMs.eps}
\includegraphics[width = \textwidth]{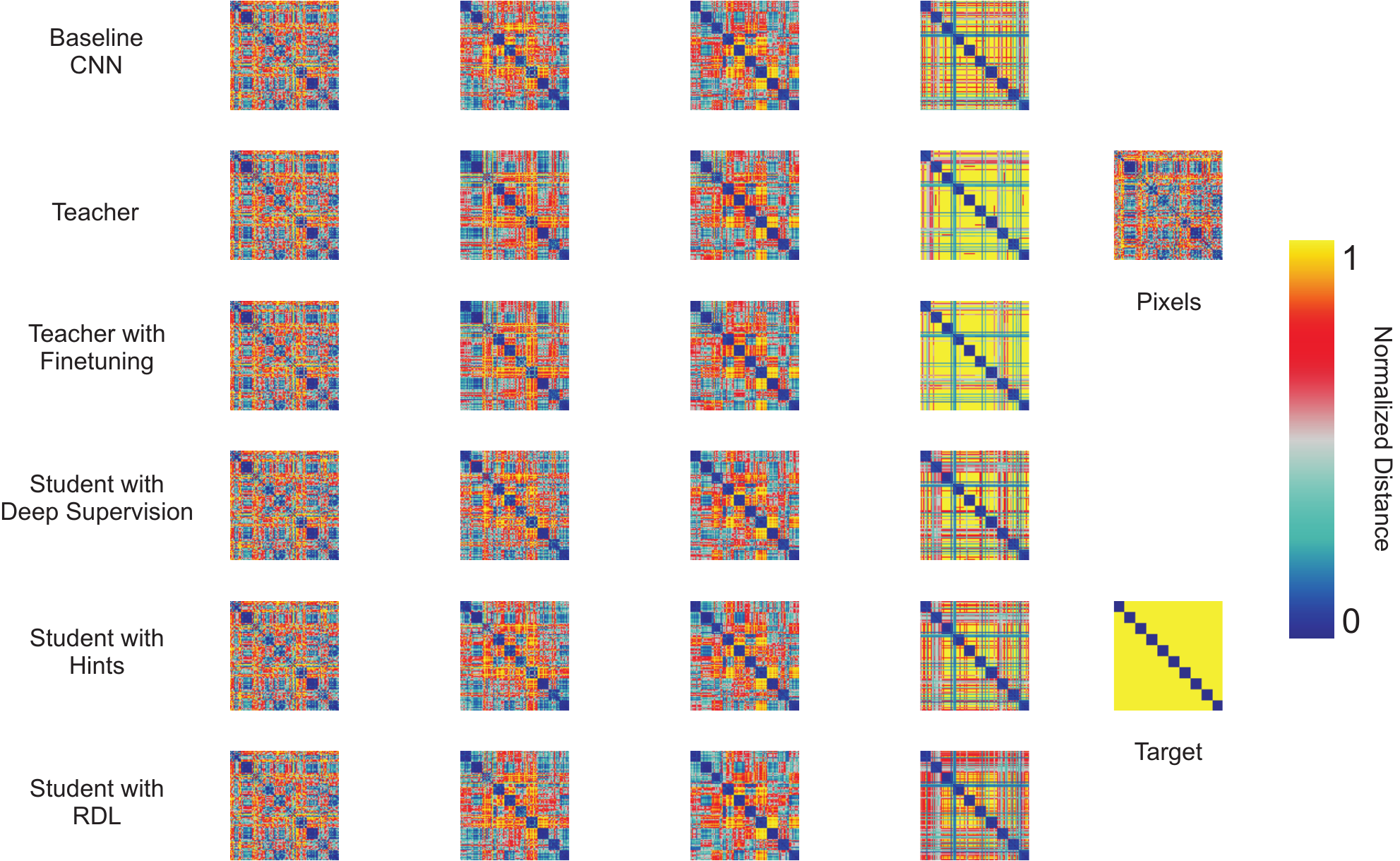}
\caption{Representational distance matrices (RDMs) using the Euclidean distance for the first and second convolutional  layers as well as the fully connected (FC) and softmax layers of the CNN tested methods, the raw pixel data, and the target labels for 10 random class exemplars from MNIST.}
\label{MNIST-RDMs}
\end{figure}

In order to further compare the trained networks, RDMs were generated for each fully trained model. Figure \ref{MNIST-RDMs} shows RDMs for 100 random test images, 10 from each class. This visualization emphasizes the class clustering as inputs are transformed from pixel space to label space. Some classes are already clustered in pixel space. For instance,  1s, 7s and 9s each have large blocks along the diagonal portion of the pixel RDM. However, by looking at the rows and columns we can see that these classes are difficult to separate from one another. After the first convolutional layer, class clustering increases, especially for the baseline CNN.  After the second convolutional layer, class clustering increases for every model and other class relationships become apparent. For instance, 3s and 5s are becoming increasingly different from other classes, but are still similar to each other. Also, 1s remain similar to many other classes. The fully connected (FC) layer leads to stronger, but not perfect, class cluster. As expected, the softmax layer leads to extremely strong class distinction. However, most of the models still view 1s as similar to other classes, as seen by the large horizontal and vertical grey stripes. The notable exception is the finetuned CNN, which had the lowest testing error.

\begin{figure}
\centering
\includegraphics[scale=.3]{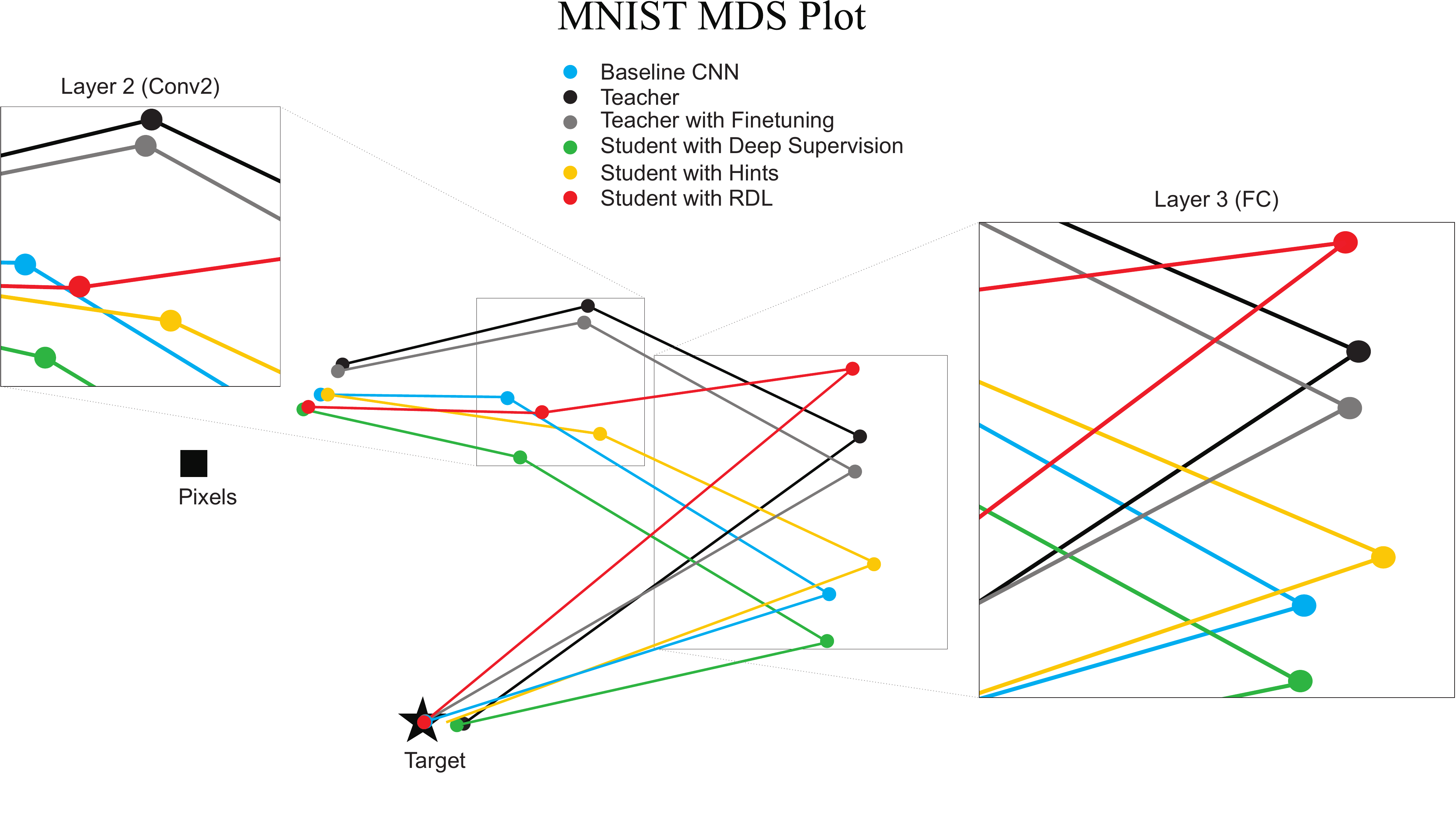}
\caption{2-D multi-dimensional scaling (MDS) visualization of the distances between the representational distance matrices (RDMs) for selected layers of the MNIST trained networks. RDMs were generated for each model using 20 bootstrapped samples of 100 images from the test set. For each sampled image set, the correlation distance between the RDMs of the different networks were calculated. These values were then averaged to generate the MDS plot.}
\label{MNIST-MDS}
\end{figure}

While viewing the RDMs directly can make certain facts about the transformations performed by the models evident, it can be hard to compare RDMs to each other by visual inspection. To better understand the relationships between the representations of the different models, we calculate the correlation distance between each pair of RDMs and use MDS to create a 2-D plot showing the relative position in representational space of the transformations learned by the various trained networks (Figure \ref{MNIST-MDS}). This allows for drawing several qualitative conclusions. As expected, the RDMs of the networks start close to the pixel-based RDM and become more similar to the target RDM the deeper the layer. The differences between the evaluated techniques can most clearly be seen at the 2nd (Conv2) and 3rd (FC) layers. As expected: (1) the network initialized with the weights of the teacher and then finetuned has the most similar RDMs to the teacher, (2) deep supervision pulls the RDMs of the student towards the target, (3) RDL pulls the RDMs of the student toward and the RDMs of the teacher, especially at 3rd layer. 

\begin{table}[b]
\centering
\caption{Test errors for MNIST and CIFAR-100 for the trained neural networks. (Note: The performance of the teacher for the CIFAR-100 classification is not shown, since it was trained on CIFAR-10 and, therefore, predicted across 10 not 100 classes, making it unable to perform the CIFAR-100 task.)}
\label{Cifar-errors}
\vskip 0.15in
\begin{small}

\begin{tabular}{l r}
\multicolumn{2}{c}{MNIST} \\
\hline
\textbf{Method} & \textbf{Error (\%)} \\
\hline
Baseline CNN    & 0.63 \\
Teacher & 0.56 \\
Teacher with Finetuning    & 0.48 \\
Student with Deep Supervision & 0.55 \\
Student with Hints & 0.56\\
Student with RDL    & 0.49 \\
\hline
\end{tabular}
\hskip 1in
\begin{tabular}{l r}
\multicolumn{2}{c}{CIFAR-100} \\
\hline
\textbf{Method} & \textbf{Error (\%)} \\
\hline
Baseline NiN    & 30.68 \\
Teacher with Finetuning    & 38.75 \\
Student with Deep Supervision & 29.46 \\
Student with Hints & 29.37 \\
Student with RDL    & 28.77 \\
\hline
\end{tabular}
\end{small}
\vskip -0.1in
\end{table}

\begin{figure}[b]
\centering

\includegraphics[scale=.3]{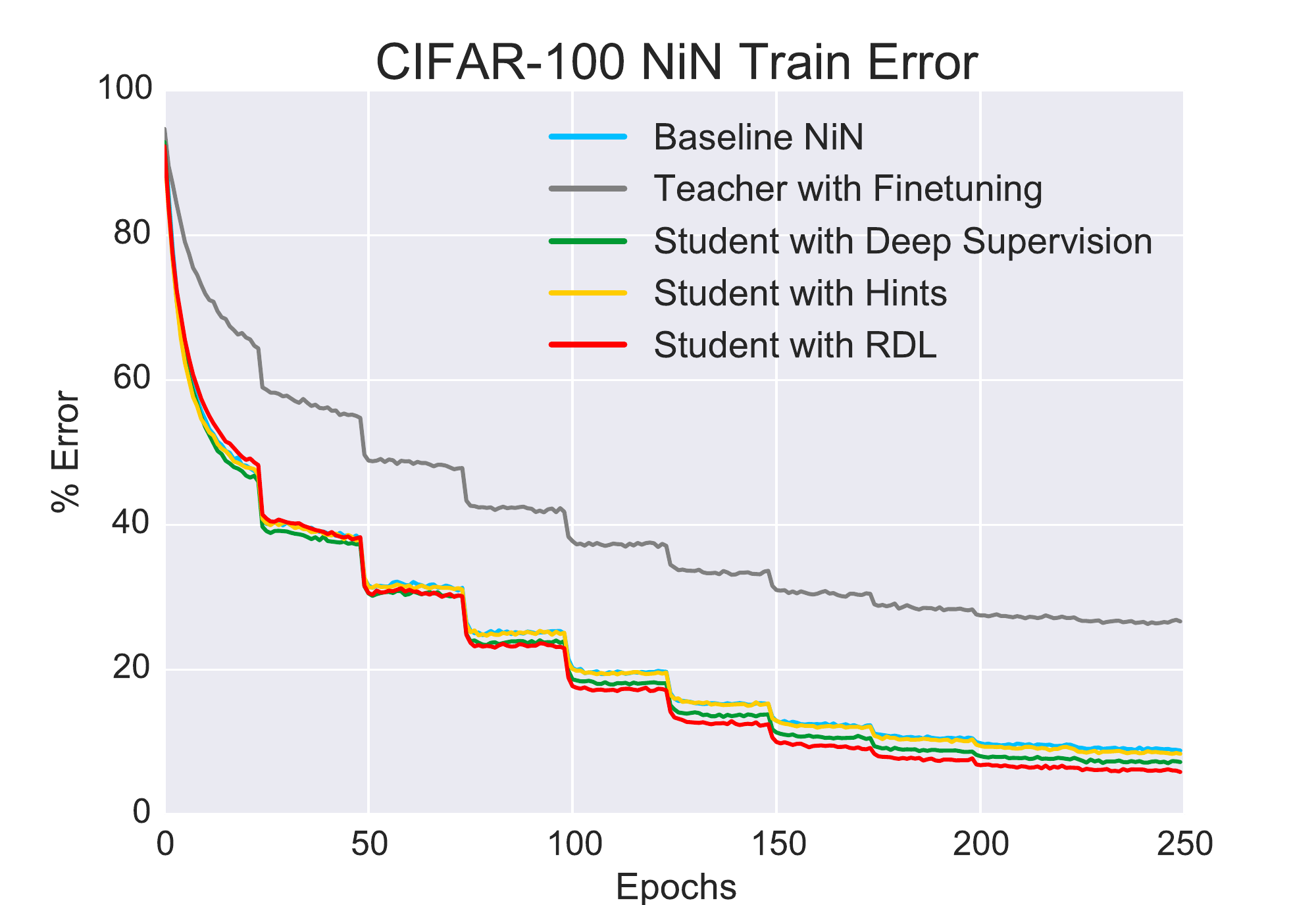}
\centering
\includegraphics[scale=.3]{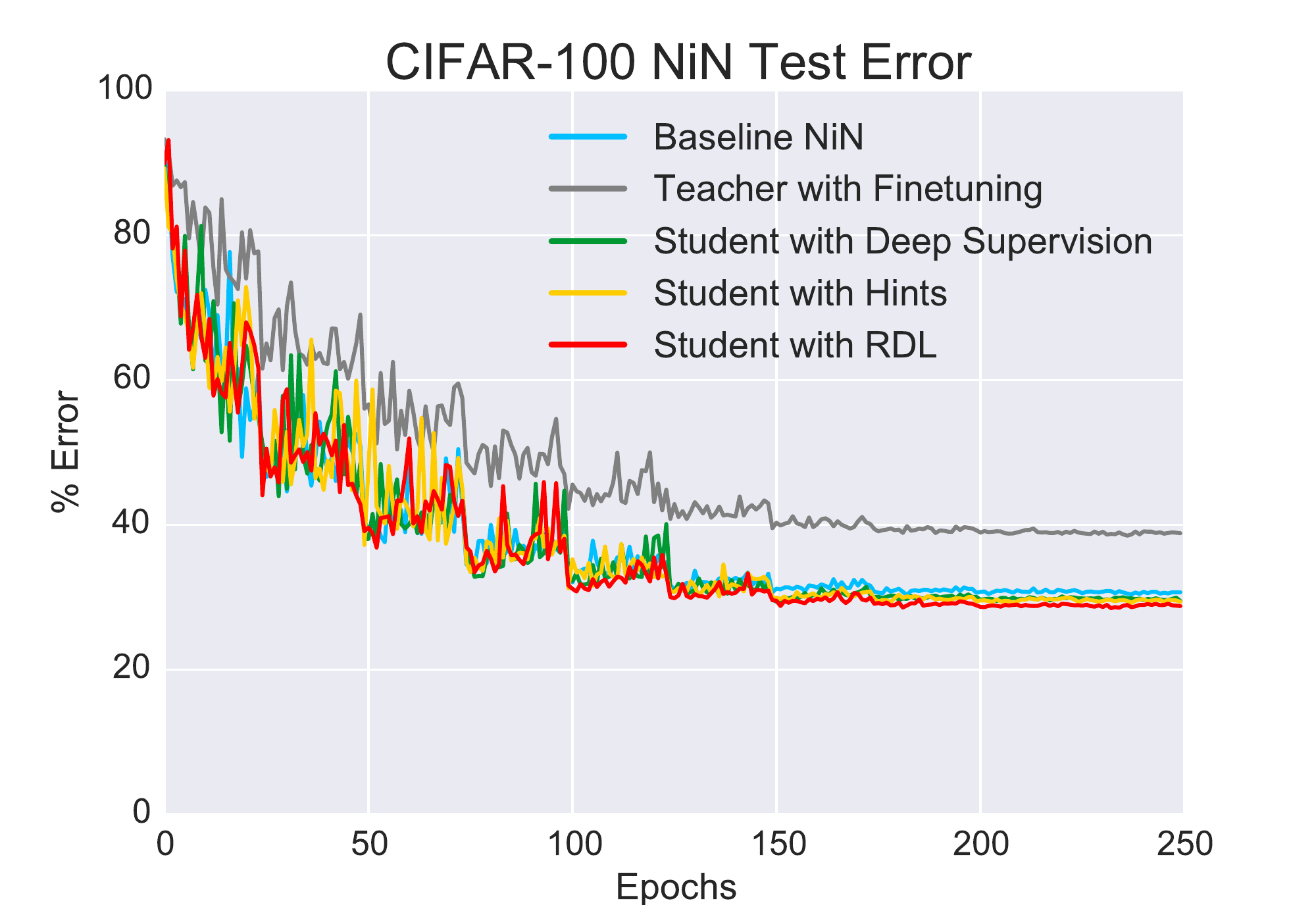}
\caption{The change in the train and test errors through time for the "Network in Network" (NiN) models trained on CIFAR-100.}
\label{fig:CIFARError}
\end{figure}

\subsection{CIFAR-100}

In order to test RDL on a more interesting problem, we performed transfer learning from CIFAR-10 to CIFAR-100. This experiment consists of transferring knowledge learned in an easier task to a harder one, something that is useful in many instances. CIFAR-100 is a dataset of 32x32 color images each containing one of one hundred objects. The dataset contains 50,000 training images and 10,000 test images. A 10,000 image subset of the training data was used as a validation set for hyper-parameter tuning. CIFAR-10 is also a dataset of 32x32 color images, but containing only ten distinct classes instead of one hundred. CIFAR-10 also contains 50,000 training images and 10,000 test images. For both datasets, the data were pre-processed using global contrast normalization. During training, random horizontal flips of the images were performed and the learning rate was halved every 25 epochs.

\begin{table}[t]
\caption{The "Network in Network" (NiN) architecture with batch-normalization (BN) \citep{ioffe2015batch} used for CIFAR-100.}
\label{CIFAR-NiN-arch}
\begin{center}
\begin{small}
\begin{tabular}{|c|c|c|c|c|c|}

\hline
Layer & Kernel Size & \# Features & Stride & Non-linearity & Other \\
\hline
Conv-1 & 5x5 & 192 & 1 & ReLU & BN \\
\hline
MLPConv-1-1 & 1x1 & 160 & 1 & ReLU & BN \\
\hline
MLPConv-1-2 & 1x1 & 96 & 1 & ReLU & BN \\
\hline
MaxPool & 3x3 & 96 & 2 & Max & -  \\
\hline
Conv-2 & 5x5 & 192 & 1 & ReLU & BN, Dropout ($p = 0.5$) \\
\hline
MLPConv-2-1 & 1x1 & 192 & 1 & ReLU & BN \\
\hline
MLPConv-2-2 & 1x1 & 192 & 1 & ReLU & BN \\
\hline
AveragePool-1 & 3x3 & 192 & 2 & - & - \\
\hline
Conv-3 & 5x5 & 192 & 1 & ReLU & BN, Dropout ($p = 0.5$) \\
\hline
MLPConv-3-1 & 1x1 & 192 & 1 & ReLU & BN \\
\hline
MLPConv-3-2 & 1x1 & 100 & 1 & ReLU & BN \\
\hline
AveragePool-2 & 8x8 & 100 & - & - & -\\
\hline
\end{tabular}
\end{small}
\end{center}
\end{table}

To evaluate using RDL with a more complex network, we used a "Network in Network" (NiN) architecture \citep{lin2013network}, which use MLPConv layers, convolutional layers that use multi-layered perception (MLP) filters instead of linear filters   (Table \ref{CIFAR-NiN-arch}). The CIFAR-10 trained teacher network had the same architecture as the baseline CIFAR-100 NiN (Table \ref{CIFAR-NiN-arch}) except with a 10-class output layer and had a testing error of 8.0\%. The DSN had linear auxiliary softmax classifiers after the first and second pooling layers and $\alpha$ was decayed as proposed in \citet{lee2014deeply}. The finetuning network's weights were initialized using those of the CIFAR-10 teacher network and a linear readout was added. The RDL network had the same architecture as the baseline CIFAR-100 network with randomly initialized weights and the addition of auxiliary error functions that used the RDMs from the CIFAR-10 teacher. For RDL, an additional linear readout was added after the last MLPConv layer since RDL does not specify that each neuron in a representation corresponds to an output class. For RDL, 2.5\% (406) of the image pairs per mini-batch of 128 images were used to calculate the RDL auxiliary errors.

\begin{figure}[t]
\centering
\includegraphics[scale=.29]{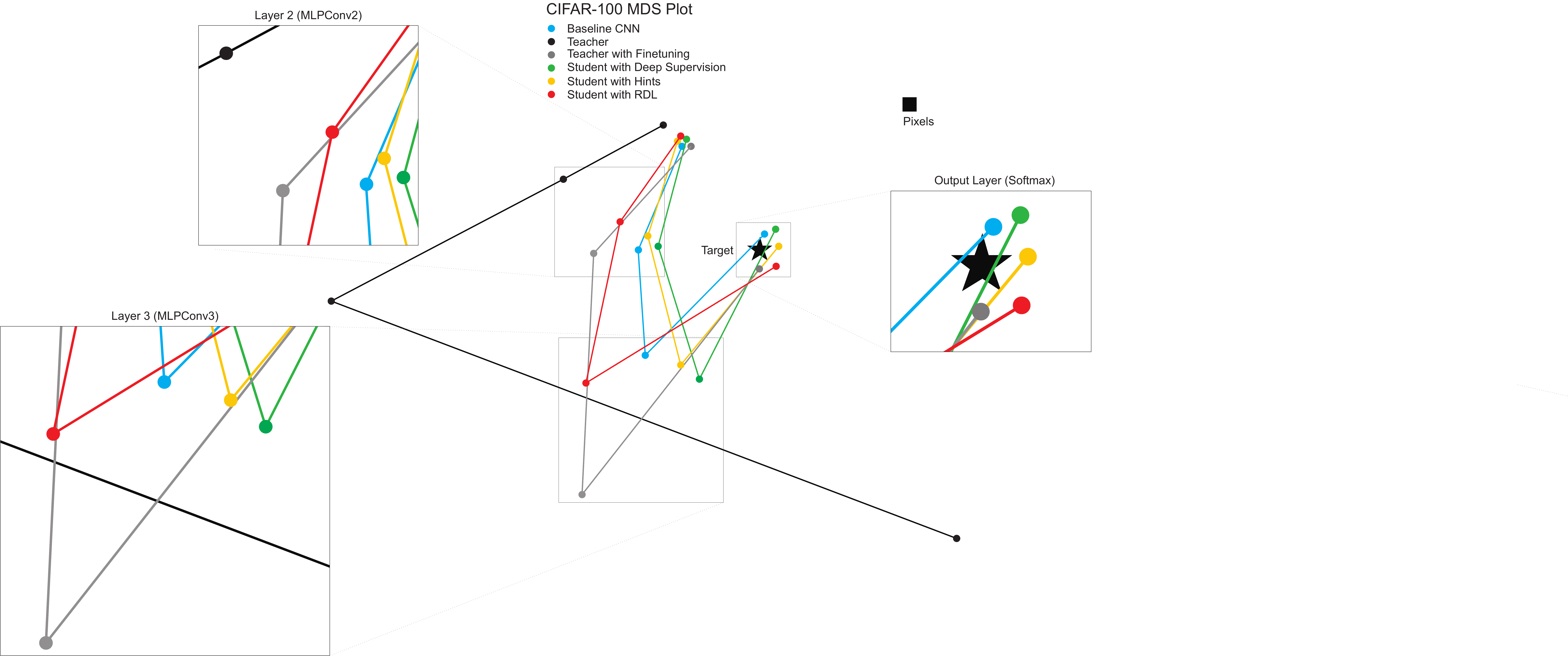}
\caption{2-D multi-dimensional scaling (MDS) visualization of the distances between the representational distance matrices (RDMs) for selected layers of the CIFAR-100 trained networks. RDMs were generated for each model using 20 bootstrapped samples of 100 images from the test set. For each sampled image set, the normalized Euclidean distance between the RDMs of the different networks were calculated. These values were then averaged to generate the MDS plot.}
\label{CIFAR-100-MDS}

\end{figure}

As in the previous experiment, the performances of the networks (Figure \ref{fig:CIFARError} and Table \ref{Cifar-errors}) were statistically compared using the McNemar test. The results are shown in Table \ref{CIFAR-100-NiN-McNemar}. Unlike in the MNIST experiment, the fine tuned network performed statistically worse than all tested methods. This is likely a combination of the weights being overspecialized for CIFAR-10 classification and the last MLPConv layer having less units. The networks that were trained with deep supervision, hints, and RDL all significantly improved upon the baseline NiN and the finetuned network. These results show that learning from RDMs can extract meaningful information from a teacher network, which leads to improved classification performance.

To investigate the relationships between the representations of the different NiN models, we calculate the correlation between each pair of RDMs and use MDS to create a 2-D plot showing the relative position in representational space of the transformations learned by the various trained networks (Figure \ref{CIFAR-100-MDS}). The MDS plots shows that: (1) the layer 2 and layer 3 RDMs of the network initialized with the weights of the teacher and then finetuned are further from the target than the other non-teacher networks, (2) deep supervision pulls the RDMs of the student towards the target, (3) despite learning a series of transformations that do not map directly to the target, the teacher contains useful information to the students' task, and (4) RDL pulls the RDMs of the student toward and the RDMs of the teacher.  This shows the ability of RDL to incorporate both the representational information from the teacher as well as from the classification task.

\begin{table}[t]
\caption{The McNemar exact test p-values for the tested "Network in Network" (NiN) models trained on CIFAR-100. Arrows indicate a significant difference ($p < 0.05$,uncorr.) and point to the better model.}
\label{CIFAR-100-NiN-McNemar}
\vskip 0.15in
\begin{small}
\begin{center}
\begin{tabular}{|c|c|c|c|c|c|}
\cline{2-6}
\multicolumn{1}{c|}{ }& Baseline & Finetuning & Deepl Supervision & Hints & RDL \\
\hline
Baseline & ---  & 0.00 $\leftarrow$ & 0.00 $\uparrow$ & 0.00 $\uparrow$ & 0.00 $\uparrow$ \\
\cline{1-1}
Finetuning & 0.00 $\uparrow$ & --- & 0.00 $\uparrow$ & 0.00 $\uparrow$ & 0.00 $\uparrow$ \\
\cline{1-1}
Deeply Supervision & 0.00 $\leftarrow$ & 0.00 $\leftarrow$ & --- & 0.86 & 0.08 \\
\cline{1-1}
Hints &  0.00 $\leftarrow$ & 0.00 $\leftarrow$ & 0.86 & --- & 0.05  \\
\cline{1-1}
RDL & 0.00 $\leftarrow$ & 0.00 $\leftarrow$ & 0.08 & 0.05 & --- \\
\hline
\end{tabular}
\end{center}
\end{small}
\end{table}

\section{Discussion}

In this paper, we proposed RDL, a technique for transferring knowledge from a teacher model to a student DNN. The representational space of the student is pulled towards that of a teacher model during training using stochastic gradient descent. This was performed by minimizing the difference between the pairwise distances between representations of two models at selected layers using auxiliary error functions. Training with RDL was shown to improve classification performance by extracting knowledge from another model trained on a similar task, while allowing architectural differences between the student and teacher. This suggests that RDL can transfer the relationships between class examples learned by the teacher. This information is not present when only constraining internal layers using class labels, as done in the deeply supervised method, since the target vectors for each class are orthogonal. In particular, RDL allows a student network to learn similar sequential transformations to those learned by a teacher network. This could be of potential use in learning transformations similar to those performed in the ventral stream. Such a model might be able to easily generate brain-like RDMs for novel stimuli. In the future, we plan on training such a model by constraining large DNNs using fMRI-based RDMs from the human ventral stream. This has the potential to help build more realistic models of the computations underlying biological visual object recognition.
\subsubsection*{Acknowledgements}

This research was funded by the Cambridge Commonwealth, European \& International Trust, the UK Medical Research Council (Program MC-A060-5PR20), and a European Research Council Starting Grant (ERC-2010-StG 261352).

\bibliographystyle{frontiersinSCNS_ENG_HUMS}
\bibliography{eccv2016}

\begin{thebibliography}{31}
\providecommand{\natexlab}[1]{#1}
\expandafter\ifx\csname urlstyle\endcsname\relax
  \providecommand{\doi}[1]{doi:\discretionary{}{}{}#1}\else
  \providecommand{\doi}{doi:\discretionary{}{}{}\begingroup
  \urlstyle{rm}\Url}\fi
\providecommand{\selectlanguage}[1]{\relax}
\providecommand{\bibAnnoteFile}[1]{%
  \IfFileExists{#1}{\begin{quotation}\noindent\textsc{Key:} #1\\
  \textsc{Annotation:}\ \input{#1}\end{quotation}}{}}
\providecommand{\bibAnnote}[2]{%
  \begin{quotation}\noindent\textsc{Key:} #1\\
  \textsc{Annotation:}\ #2\end{quotation}}

\bibitem[{Ba and Caruana(2014)}]{ba2014deep}
Ba, J. and Caruana, R. (2014).
\newblock Do deep nets really need to be deep?
\newblock In \emph{Advances in Neural Information Processing Systems}.
  2654--2662
\bibAnnoteFile{ba2014deep}

\bibitem[{Belkin and Niyogi(2003)}]{belkin2003laplacian}
Belkin, M. and Niyogi, P. (2003).
\newblock Laplacian eigenmaps for dimensionality reduction and data
  representation.
\newblock \emph{Neural computation} 15, 1373--1396
\bibAnnoteFile{belkin2003laplacian}

\bibitem[{Bengio(2012)}]{bengio2012deep}
Bengio, Y. (2012).
\newblock Deep learning of representations for unsupervised and transfer
  learning.
\newblock \emph{Unsupervised and Transfer Learning Challenges in Machine
  Learning} 7, 19
\bibAnnoteFile{bengio2012deep}

\bibitem[{Bucilua et~al.(2006)Bucilua, Caruana, and
  Niculescu-Mizil}]{bucilua2006model}
Bucilua, C., Caruana, R., and Niculescu-Mizil, A. (2006).
\newblock Model compression.
\newblock In \emph{Proceedings of the 12th ACM SIGKDD international conference
  on Knowledge discovery and data mining} (ACM), 535--541
\bibAnnoteFile{bucilua2006model}

\bibitem[{Collobert et~al.(2011)Collobert, Kavukcuoglu, and
  Farabet}]{collobert2011torch7}
Collobert, R., Kavukcuoglu, K., and Farabet, C. (2011).
\newblock Torch7: A matlab-like environment for machine learning.
\newblock In \emph{BigLearn, NIPS Workshop}. EPFL-CONF-192376
\bibAnnoteFile{collobert2011torch7}

\bibitem[{Deng et~al.(2013)Deng, Hinton, and Kingsbury}]{deng2013new}
Deng, L., Hinton, G., and Kingsbury, B. (2013).
\newblock New types of deep neural network learning for speech recognition and
  related applications: An overview.
\newblock In \emph{Acoustics, Speech and Signal Processing (ICASSP), 2013 IEEE
  International Conference on} (IEEE), 8599--8603
\bibAnnoteFile{deng2013new}

\bibitem[{Edwards(1948)}]{edwards1948note}
Edwards, A.~L. (1948).
\newblock Note on the “correction for continuity” in testing the
  significance of the difference between correlated proportions.
\newblock \emph{Psychometrika} 13, 185--187
\bibAnnoteFile{edwards1948note}

\bibitem[{G{\"u}{\c{c}}l{\"u} and van Gerven(2014)}]{gucclu2014deep}
G{\"u}{\c{c}}l{\"u}, U. and van Gerven, M.~A. (2014).
\newblock Deep neural networks reveal a gradient in the complexity of neural
  representations across the brain's ventral visual pathway.
\newblock \emph{arXiv preprint arXiv:1411.6422}
\bibAnnoteFile{gucclu2014deep}

\bibitem[{Hinton et~al.(2015)Hinton, Vinyals, and Dean}]{hinton2015distilling}
Hinton, G., Vinyals, O., and Dean, J. (2015).
\newblock Distilling the knowledge in a neural network.
\newblock \emph{arXiv preprint arXiv:1503.02531}
\bibAnnoteFile{hinton2015distilling}

\bibitem[{Ioffe and Szegedy(2015)}]{ioffe2015batch}
Ioffe, S. and Szegedy, C. (2015).
\newblock Batch normalization: Accelerating deep network training by reducing
  internal covariate shift.
\newblock \emph{arXiv preprint arXiv:1502.03167}
\bibAnnoteFile{ioffe2015batch}

\bibitem[{Khaligh-Razavi and Kriegeskorte(2014)}]{khaligh2014deep}
Khaligh-Razavi, S.-M. and Kriegeskorte, N. (2014).
\newblock Deep supervised, but not unsupervised, models may explain it cortical
  representation.
\newblock \emph{PLoS Comput Biol} 10, e1003915
\bibAnnoteFile{khaligh2014deep}

\bibitem[{Kriegeskorte et~al.(2008)Kriegeskorte, Mur, and
  Bandettini}]{kriegeskorte2008representational}
Kriegeskorte, N., Mur, M., and Bandettini, P. (2008).
\newblock Representational similarity analysis--connecting the branches of
  systems neuroscience.
\newblock \emph{Frontiers in systems neuroscience} 2
\bibAnnoteFile{kriegeskorte2008representational}

\bibitem[{Krizhevsky et~al.(2012)Krizhevsky, Sutskever, and
  Hinton}]{krizhevsky2012imagenet}
Krizhevsky, A., Sutskever, I., and Hinton, G.~E. (2012).
\newblock Imagenet classification with deep convolutional neural networks.
\newblock In \emph{Advances in neural information processing systems}.
  1097--1105
\bibAnnoteFile{krizhevsky2012imagenet}

\bibitem[{Kruskal(1964)}]{kruskal1964multidimensional}
Kruskal, J.~B. (1964).
\newblock Multidimensional scaling by optimizing goodness of fit to a nonmetric
  hypothesis.
\newblock \emph{Psychometrika} 29, 1--27
\bibAnnoteFile{kruskal1964multidimensional}

\bibitem[{LeCun et~al.(1998)LeCun, Bottou, Bengio, and
  Haffner}]{lecun1998gradient}
LeCun, Y., Bottou, L., Bengio, Y., and Haffner, P. (1998).
\newblock Gradient-based learning applied to document recognition.
\newblock \emph{Proceedings of the IEEE} 86, 2278--2324
\bibAnnoteFile{lecun1998gradient}

\bibitem[{Lee et~al.(2014)Lee, Xie, Gallagher, Zhang, and Tu}]{lee2014deeply}
Lee, C.-Y., Xie, S., Gallagher, P., Zhang, Z., and Tu, Z. (2014).
\newblock Deeply-supervised nets.
\newblock \emph{arXiv preprint arXiv:1409.5185}
\bibAnnoteFile{lee2014deeply}

\bibitem[{Lin et~al.(2013)Lin, Chen, and Yan}]{lin2013network}
Lin, M., Chen, Q., and Yan, S. (2013).
\newblock Network in network.
\newblock \emph{arXiv preprint arXiv:1312.4400}
\bibAnnoteFile{lin2013network}

\bibitem[{Loosli et~al.(2007)Loosli, Canu, and Bottou}]{loosli2006training}
Loosli, G., Canu, S., and Bottou, L. (2007).
\newblock Training invariant support vector machines using selective sampling.
\newblock In \emph{Large Scale Kernel Machines}, eds. L.~Bottou, O.~Chapelle,
  D.~{DeCoste}, and J.~Weston (Cambridge, MA.: MIT Press). 301--320
\bibAnnoteFile{loosli2006training}

\bibitem[{Naselaris et~al.(2011)Naselaris, Kay, Nishimoto, and
  Gallant}]{naselaris2011encoding}
Naselaris, T., Kay, K.~N., Nishimoto, S., and Gallant, J.~L. (2011).
\newblock Encoding and decoding in fmri.
\newblock \emph{Neuroimage} 56, 400--410
\bibAnnoteFile{naselaris2011encoding}

\bibitem[{Nili et~al.(2014)Nili, Wingfield, Walther, Su, Marslen-Wilson, and
  Kriegeskorte}]{nili2014toolbox}
Nili, H., Wingfield, C., Walther, A., Su, L., Marslen-Wilson, W., and
  Kriegeskorte, N. (2014).
\newblock A toolbox for representational similarity analysis.
\newblock \emph{PLoS Comput. Biol} 10, e1003553
\bibAnnoteFile{nili2014toolbox}

\bibitem[{Romero et~al.(2014)Romero, Ballas, Kahou, Chassang, Gatta, and
  Bengio}]{romero2014fitnets}
Romero, A., Ballas, N., Kahou, S.~E., Chassang, A., Gatta, C., and Bengio, Y.
  (2014).
\newblock Fitnets: Hints for thin deep nets.
\newblock \emph{arXiv preprint arXiv:1412.6550}
\bibAnnoteFile{romero2014fitnets}

\bibitem[{Russakovsky et~al.(2014)Russakovsky, Deng, Su, Krause, Satheesh, Ma
  et~al.}]{russakovsky2014imagenet}
Russakovsky, O., Deng, J., Su, H., Krause, J., Satheesh, S., Ma, S., et~al.
  (2014).
\newblock Imagenet large scale visual recognition challenge.
\newblock \emph{International Journal of Computer Vision} , 1--42
\bibAnnoteFile{russakovsky2014imagenet}

\bibitem[{Szegedy et~al.(2014)Szegedy, Liu, Jia, Sermanet, Reed, Anguelov
  et~al.}]{szegedy2014going}
Szegedy, C., Liu, W., Jia, Y., Sermanet, P., Reed, S., Anguelov, D., et~al.
  (2014).
\newblock Going deeper with convolutions.
\newblock \emph{arXiv preprint arXiv:1409.4842}
\bibAnnoteFile{szegedy2014going}

\bibitem[{Wang et~al.(2015)Wang, Lee, Tu, and Lazebnik}]{wang2015training}
Wang, L., Lee, C.-Y., Tu, Z., and Lazebnik, S. (2015).
\newblock Training deeper convolutional networks with deep supervision.
\newblock \emph{arXiv preprint arXiv:1505.02496}
\bibAnnoteFile{wang2015training}

\bibitem[{Weston et~al.(2012)Weston, Ratle, Mobahi, and
  Collobert}]{weston2012deep}
Weston, J., Ratle, F., Mobahi, H., and Collobert, R. (2012).
\newblock Deep learning via semi-supervised embedding.
\newblock In \emph{Neural Networks: Tricks of the Trade} (Springer). 639--655
\bibAnnoteFile{weston2012deep}

\bibitem[{Yamins and DiCarlo(2016)}]{yamins2016using}
Yamins, D.~L. and DiCarlo, J.~J. (2016).
\newblock Using goal-driven deep learning models to understand sensory cortex.
\newblock \emph{Nature neuroscience} 19, 356--365
\bibAnnoteFile{yamins2016using}

\bibitem[{Yamins et~al.(2014)Yamins, Hong, Cadieu, Solomon, Seibert, and
  DiCarlo}]{yamins2014performance}
Yamins, D.~L., Hong, H., Cadieu, C.~F., Solomon, E.~A., Seibert, D., and
  DiCarlo, J.~J. (2014).
\newblock Performance-optimized hierarchical models predict neural responses in
  higher visual cortex.
\newblock \emph{Proceedings of the National Academy of Sciences} 111,
  8619--8624
\bibAnnoteFile{yamins2014performance}

\bibitem[{Yosinski et~al.(2014)Yosinski, Clune, Bengio, and
  Lipson}]{yosinski2014transferable}
Yosinski, J., Clune, J., Bengio, Y., and Lipson, H. (2014).
\newblock How transferable are features in deep neural networks?
\newblock In \emph{Advances in Neural Information Processing Systems}.
  3320--3328
\bibAnnoteFile{yosinski2014transferable}

\bibitem[{Yosinski et~al.(2015)Yosinski, Clune, Nguyen, Fuchs, and
  Lipson}]{yosinski2015understanding}
Yosinski, J., Clune, J., Nguyen, A., Fuchs, T., and Lipson, H. (2015).
\newblock Understanding neural networks through deep visualization.
\newblock \emph{arXiv preprint arXiv:1506.06579}
\bibAnnoteFile{yosinski2015understanding}

\bibitem[{Zeiler and Fergus(2014)}]{zeiler2014visualizing}
Zeiler, M.~D. and Fergus, R. (2014).
\newblock Visualizing and understanding convolutional networks.
\newblock In \emph{Computer Vision--ECCV 2014} (Springer). 818--833
\bibAnnoteFile{zeiler2014visualizing}

\bibitem[{Zhou et~al.(2014)Zhou, Lapedriza, Xiao, Torralba, and
  Oliva}]{zhou2014learning}
Zhou, B., Lapedriza, A., Xiao, J., Torralba, A., and Oliva, A. (2014).
\newblock Learning deep features for scene recognition using places database.
\newblock In \emph{Advances in Neural Information Processing Systems}. 487--495
\bibAnnoteFile{zhou2014learning}

\end{thebibliography}

\end{document}